\documentclass[runningheads,a4paper]{llncs}

\usepackage{amssymb}
\usepackage{graphicx}

\usepackage{url}
\urldef{\mailsa}\path|{alfred.hofmann, ursula.barth, ingrid.haas, frank.holzwarth,|
\urldef{\mailsb}\path|anna.kramer, leonie.kunz, christine.reiss, nicole.sator,|
\urldef{\mailsc}\path|erika.siebert-cole, peter.strasser, lncs}@springer.com|    
\newcommand{\keywords}[1]{\par\addvspace\baselineskip
\noindent\keywordname\enspace\ignorespaces#1}
\newcommand*{\affaddr}[1]{#1} 
\newcommand*{\affmark}[1][*]{\textsuperscript{#1}}

\usepackage{subfig}
\usepackage{amsmath}
\usepackage{multirow}

\begin{document}

\mainmatter  

\title{Deep Vessel Segmentation\\By Learning Graphical Connectivity}

\titlerunning{Deep Vessel Segmentation By Learning Graphical Connectivity}

%
%
\author{
Seung Yeon Shin\affmark[1\dag] \and Soochahn Lee\affmark[2] \and Il Dong Yun\affmark[3] \and Kyoung Mu Lee\affmark[1] 
}
\authorrunning{S.Y. Shin et al.}
%

\institute{
\affaddr{\affmark[1]Dept. ECE, ASRI, Seoul Nat'l Univ.},
\affaddr{\affmark[2]Dept. Electronic Eng., Soonchunhyang Univ.},\\
\affaddr{\affmark[3]Div. Comp. \& Elec. Sys. Eng., Hankuk Univ. of Foreign Studies}\\
\email{\affmark[\dag]syshin@snu.ac.kr}
}

%
%

\toctitle{Deep Vessel Segmentation By Learning Graphical Connectivity}
\tocauthor{S.Y. Shin et al.}
\maketitle

\begin{abstract}
	
We propose a novel deep-learning-based system for vessel segmentation. Existing methods using CNNs have mostly relied on local appearances learned on the regular image grid, without considering the graphical structure of vessel shape. To address this, we incorporate a graph convolutional network into a unified CNN architecture, where the final segmentation is inferred by combining the different types of features. The proposed method can be applied to expand any type of CNN-based vessel segmentation method to enhance the performance. Experiments show that the proposed method outperforms the current state-of-the-art methods on two retinal image datasets as well as a coronary artery X-ray angiography dataset. 

\end{abstract}

\section{Introduction}

Observation of blood vessels is crucial in the diagnosis and intervention of many diseases. Clinicians have mainly relied on manual inspections, which can be inaccurate and time-consuming. Over the years, the demand for efficiency has led to the development of numerous methods for automatic vessel segmentation.

Most methods are based on image-processing~\cite{soares06}, optimization~\cite{orlando14,shin16}, learning~\cite{becker13,sironi15}, or their combination. Many optimization methods, including energy minimization methods on a Markov random field~\cite{orlando14,shin16}, aim to determine the best global graph structure based on applied prior knowledge. However, prior knowledge often consists of only simple rules, limiting the modeling capacity. More complex distributions can be modeled using learning-based methods such as boosting~\cite{becker13} or regression~\cite{sironi15}. However, due to model complexity, only local appearances are mostly learned. Even with deep learning methods based on convolutional neural networks (CNN)~\cite{ganin14,fu16,maninis16} this limitation persists.

\begin{figure}
	\centering
	\includegraphics[width = 1\linewidth]{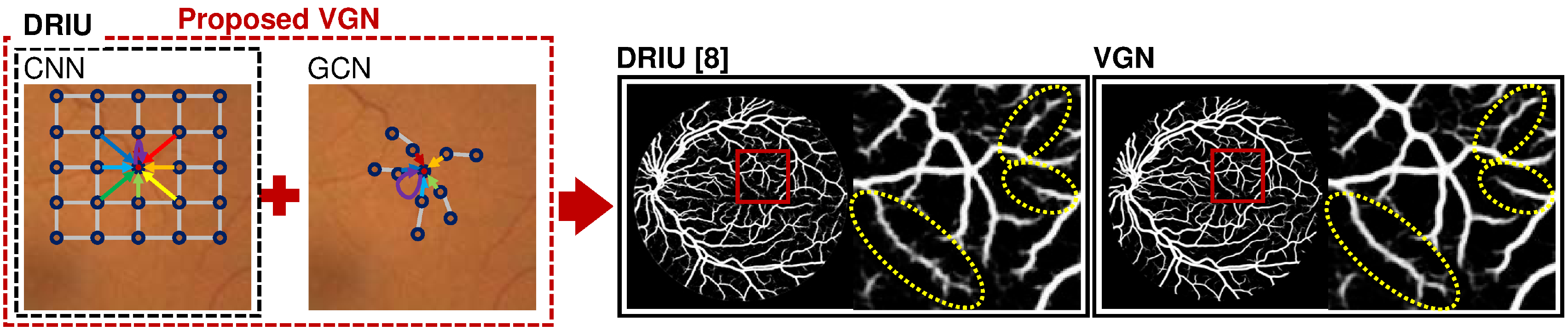}
	\caption{Motivation of the proposed method. Learning about the strong relationship that exists between neighborhoods is not guaranteed in existing CNN-based vessel segmentation methods. The proposed {\em vessel graph network} (VGN) utilizes a GCN together with a CNN to address this issue. All figures best viewed in color.}
	\label{fig:idea}
\end{figure}

\begin{figure}
	\centering
	\includegraphics[width = 0.99\linewidth]{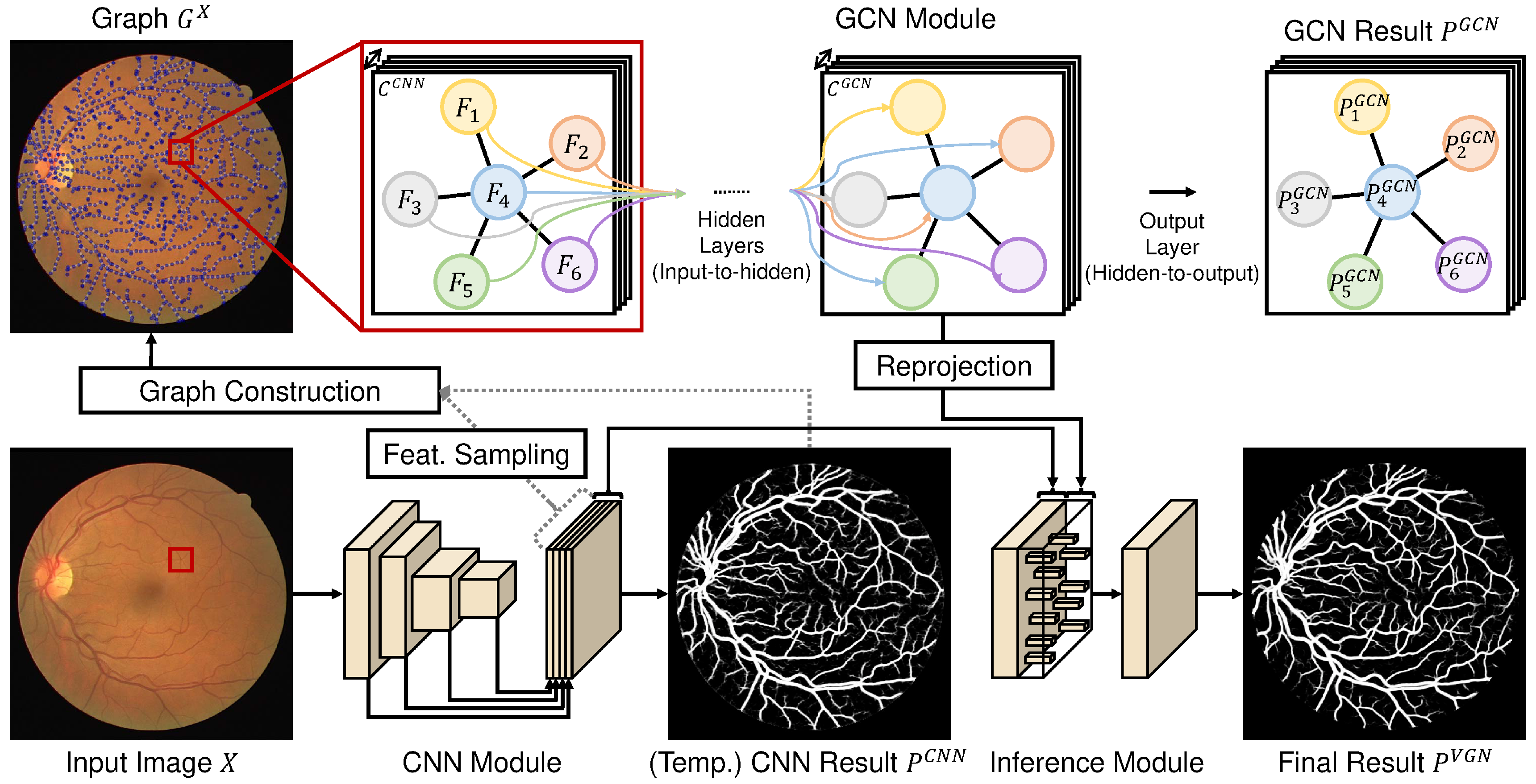}
	\caption{Overall network architecture of VGN comprising the CNN, graph convolutional network, and inference modules. Refer to text for more details.}
	\label{fig:network}
\end{figure}

Thus, we present a novel CNN architecture, the {\em vessel graph network} (VGN), that jointly learns the global structure of vessel shape together with local appearances, as shown in Fig.~\ref{fig:idea}. The VGN comprises three components, i) a CNN module for generating pixelwise features and vessel probabilities, ii) a graph convolutional network (GCN)~\cite{kipf17} module to extract features which reflect the connectivity of neighboring vertices, and iii) an inference module to produce the final segmentation. The input graph for the GCN is generated in an additional graph construction module. The network architecture is described in Fig.~\ref{fig:network}. 

The technical contributions are as follows. 1) Our work is the first, to our knowledge, method to apply GCN to learn graphical structure of blood vessels. 2) The VGN combines the GCN within a CNN structure to jointly learn both local appearance and global structure. 3) The VGN structure is widely applicable since it can be combined with any CNN-based method. 4) When extending CNN-based methods to VGN, performance is highly likely to improve, with no risk of degradation. This is because, should the GCN have no positive impact, the VGN will be trained to perform inference only from the CNN features. 5) We perform comparative evaluations on two retinal image datasets and a coronary X-ray angiography dataset, showing that the VGN outperforms current state-of-the-art (SotA) methods. 

\section{Methods}

\subsection{Overview of Network Architecture}\label{architecture}

The CNN module learns features, on the regular image grid of size $h \times w$, to infer pixelwise vessel probability $P^{CNN}=\{p^{CNN}(x_{i})\}^{h \times w}_{i=1}$ of input image $X=\{x_{i}\}^{h \times w}_{i=1}$. The GCN module learns features for vertices on an irregular graph constructed from points sampled from vessel centerlines of an initial segmentation $P^{CNN}$. Due to their interaction in the GCN, the hidden representation of each vertex $v_{j}$ reflects the likelihood of being a vessel, given the likelihood of neighboring vertices, respectively. For instance, when a vertex is surrounded by vessel vertices, it will become more likely to be labeled a vessel based on its GCN features. The combined CNN and GCN features are given to the inference module to compute the final vessel probability map $P^{VGN}=\{p^{VGN}(x_{i})\}^{h \times w}_{i=1}$.

As the CNN module, we adopt the network of DRIU (deep retinal image understanding)~\cite{maninis16}, based on the VGG-16 network~\cite{simonyan14}, due to its SotA performance. We note that any other CNN-based vessel segmentation method can be used. In DRIU, a vessel probability map is inferred from concatenated multi-scale features from the VGG-16. Before the concatenation, feature maps are resized to have identical scale. In our VGN, we adopt the pixelwise cross entropy loss $L_{CNN}(X)$ for this CNN module. Please refer to \cite{maninis16} for more details.

\subsection{Graph Convolutional Network Module}\label{gcn_module}

A graph must be constructed and given as input for both training and testing of the GCN module. We assume a CNN has been pretrained to generate $P^{CNN}$, on which the following is performed: 1) thresholding, 2) skeletonization by morphological thinning, 3) vertex generation by equidistant sampling, with distance $\delta$, on the skeleton together with skeletal junctions and endpoints, and 4) edge generation between vertices based on the skeletal connectivity or geodesic distances on the vessel probability map.


We denote the constructed graph from the image $X$ as $G^{X}(V,E)$, where $V=\{v_j\}_{j=1}^{N}$ and $E$ are sets of vertices and edges, respectively. $N$ is the number of vertices. Input feature vector $F_{j}$ for each vertex is sampled from the intermediate feature map generated from the CNN at the pixel coordinate of each vertex. The matrix of all $F_{j}$'s is denoted as $F\in\mathbb{R}^{N \times C^{CNN}}$, where $C^{CNN}$ is the feature dimension of the CNN. While the existence and weight of the edge $e_{ij}$ between the $i_{th}$ and $j_{th}$ vertices can be defined in various ways, we empirically use simple binary values based on nearest neighbor connectivity on the skeleton. The adjacency matrix defined by all $e_{ij}$'s is denoted as $A\in\mathbb{R}^{N \times N}$.

The GCN operates on the extracted graph as a vertex classifier into vessel or non-vessel. It is defined as a two-layer feed-forward model formulated as:

\begin{equation}
\label{eq:GCN_equation}
P^{GCN}(V)=f(F,A)=\sigma(\hat{A}\ ReLU(\hat{A}FW^{(0)})W^{(1)}),  
\end{equation}   
with $\widetilde{A}=A+I_N$, ${\widetilde{D}}_{rc}=\sum_{c}{\widetilde{A}}_{rc}$, and $\hat{A}={\widetilde{D}}^{-\frac{1}{2}}\widetilde{A}{\widetilde{D}}^{-\frac{1}{2}}$. $P^{GCN}(V)$ is a vessel probability vector for all vertices, which is calculated by applying the sigmoid function $\sigma$ on the final features. $W^{(0)}\in\mathbb{R}^{C^{CNN} \times C^{GCN}}$ and $W^{(1)}\in\mathbb{R}^{C^{GCN} \times 1}$ are trainable weight matrices, where $C^{GCN}$ is the number of hidden units in the GCN. More layers showed no improvement in our experiments as reported in \cite{kipf17}. For training the GCN, we use a mean of the vertex-wise cross entropy loss defined as:

\begin{equation}
\label{eq:GCN_loss}
L_{GCN}(G^{X}) = \ -\frac{1}{N}\sum_{j\in V}\sum_{l\in\{BG,FG\}}{p_l^*(x_{v2p(v_j)})log\ p_l^{GCN}(v_j)},
\end{equation}
where $v2p(v_j)$ returns the pixel index corresponding to $v_j$.  $p_l^*(x_{v2p(v_j)})$ and $p_l^{GCN}(v_j)$ are the GT label and the vessel probability predicted for $v_j$ by the GCN, respectively. $BG$ and $FG$ represent the back/foreground classes. 

\subsection{Inference Module}\label{infer_module}

To conduct inference on the combined features from the CNN and GCN modules, the spatial dimensions of the features must be normalized. Thus, we reproject the $N$ number of GCN hidden features, only sparsely present on $v_{j}$'s to the corresponding pixel coordinate in the pixelwise regular grid to coincide with the CNN features. The combined features are represented in a tensor of dimension $h \times w \times (C^{CNN}+C^{GCN})$. Since all intermediate layers are ReLU-activated in the VGN, the zero-padding on the non-vertex pixels of the GCN features can be interpreted as those are not activated.

$P^{GVN}$ is produced by applying multiple convolution layers on the combined feature tensor. To spread the sparse activations out over the whole image region, the number of the layers and the kernel sizes are determined according to the vertex sampling distance $\delta$. We empirically adopted a plain architecture composed of five convolution layers, all with kernel size 3$\times$3. For training, we again use a mean of the pixelwise cross entropy loss defined as:

\begin{equation}
\label{eq:INFER_loss}
L_{INFER}(X) =\ -\frac{1}{|X|}\sum_{i}\alpha\sum_{l\in\{BG,FG\}}{p_l^*(x_i)log\ p_l^{GVN}(x_i)},
\end{equation}
where $p_l^{GVN}(x_i)$ is the prediction for each pixel $x_i$ from the inference module. The weights for class-balancing are omitted for brevity. Here, $\alpha= 
\begin{cases}
	\delta^2, & \text{if } p2v(x_i) \in V\\
	1, & \text{otherwise}
\end{cases}$ is adopted to prevent trivial solutions which can be inferred only using the CNN features. $p2v$ is the inverse operator of $v2p$. $\delta^2$ is used since a single pixel is selected as a graph vertex among approximately $\delta^2$ pixels.

\subsection{Network Training}\label{net_training}

We adopt a sequential training scheme composed of an initial pretraining of the CNN followed by joint training, including fine-tuning of the CNN module, of the whole VGN. The $P^{CNN}$ inferred from the pretrained CNN is used to construct the training graphs as described in Section \ref{gcn_module}. To maintain efficiency, graph construction is performed only at each $K_{gc}$ training iterations. Compared to when pretraining the CNN, the VGN takes image $X$ as well as $G^X$ for joint training. With the assumption of the accuracy of the graph $G^X$ constructed from the pretrained CNN module, the proposed network learns the graphical vessel structure while fine-tuning the CNN module end-to-end. The total loss function used for the VGN is defined as:

\begin{equation}
\label{eq:total_loss}
L_{total}(X) = L_{CNN}(X) + L_{GCN}(G^{X}) + L_{INFER}(X).
\end{equation}

When testing the VGN, CNN module feature generation and inference, graph construction, GCN feature generation, and final VGN inference are performed sequentially for each image to generate the final segmentation. 

\section{Experimental Results}

\textbf{Evaluation Details}:
We experiment on two retinal image datasets, namely DRIVE~\cite{staal04} and STARE~\cite{hoover00}, and a coronary artery X-ray angiography dataset (CA-XRA). For the DRIVE and STARE sets, which respectively comprises 40 and 20 images, we followed \cite{maninis16} for training/test set splits, and human performance measurement using second observer results. CA-XRA was acquired in our cooperating hospital and comprises 3,137 image frames from 85 XRA sequences. All sequences were acquired at 512 $\times$ 512 resolution, 8 bit depth, and 15 fps frame rate. We treated each frame as an independent image, without use of temporal information. Frames of the first 80, and the last 5 sequences were assigned as training and test sets comprising 2,958 and 179 images, respectively.

Since the authors have not made their training code publicly available, our own implementation of the DRIU~\cite{maninis16} was used as the CNN module. We provide a comparison between results from our implementation and those reported in \cite{maninis16} for reference. CNN architectures are identical to the original DRIU for the retinal image datasets ($C^{CNN}$=64=16$\times$4), but slightly modified to include all five stages of VGG-16 ($C^{CNN}$=80=16$\times$5) to handle the wider variance of vessel width in CA-XRA images.
The numbers of hidden units in the GCN were set to be equal to that of the CNN, as $C^{GCN}$=64 for DRIVE/STARE, and $C^{GCN}$=80 for CA-XRA. In the inference module, feature depth is halved in layer $1$, kept constant in layer $2-4$, and reduced to 1 in the final layer $5$.

The details on CNN pretraining mostly followed that of the original DRIU~\cite{maninis16}, but slightly modified the loss function from the sum of pixelwise cross entropy to the mean, and modified the learning rate accordingly. For training the VGN, We use stochastic gradient descent with momentum of 0.9 and a weight decay of 0.0005. The initial learning rates of the pratrained CNN and the remaining modules are $0$ and $10^{-2}$ for DRIVE/STARE, $10^{-6}$ and $10^{-3}$ for CA-XRA. We found that not fine-tuning the pretrained CNN shows better results for DRIVE/STARE due to its small number of training images. The learning rate is scheduled to gradually decrease. 50,000 iterations with mini-batch size 1 were run for DRIVE/STARE, while 100,000 iterations with mini-batch size 5 were run for CA-XRA. We applied horizontal flipping, random brightness and contrast adjustment for data augmentation. Precomputed graphs are flipped accordingly. 

The graph update period $K_{gc}$ is set as 10,000 and 20,000 for retinal and CA-XRA datasets, respectively. The vertex sampling rate $\delta$ is fixed to 10. We use several thresholds for vessel probability maps during the graph construction, to simulate variability of training graph data. A randomly selected graph is used at each iteration.
In test time, the thresholds are all used and the average vessel map is given as the final output.

\begin{figure*}[t]
	\centering
	\begin{minipage}{1\linewidth}
		\subfloat{\includegraphics[width = 0.5\linewidth]{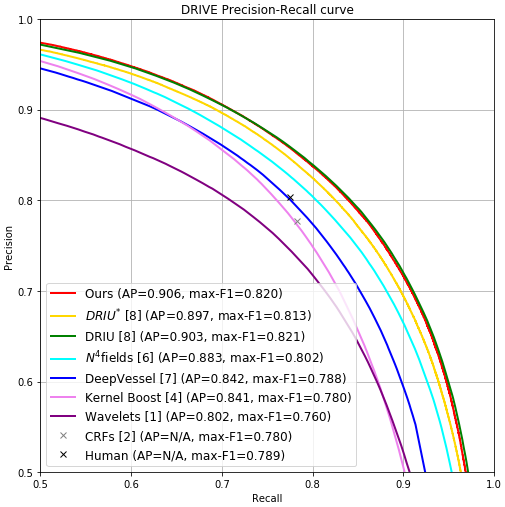}}
		\subfloat{\includegraphics[width = 0.5\linewidth]{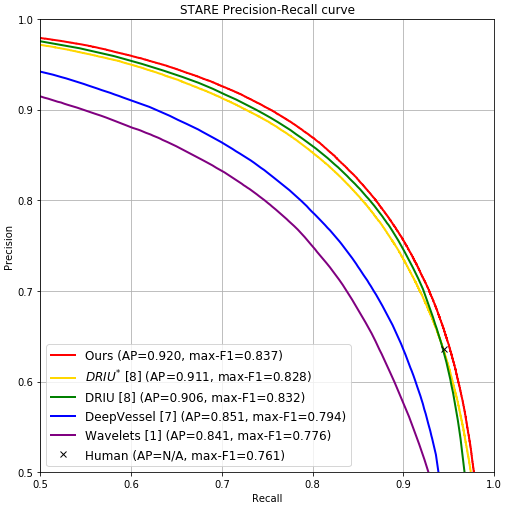}}
	\end{minipage}
	\caption{Precision recall curves, average precisions (AP), and max F1 scores of the proposed VGN and comparable methods on the DRIVE and STARE dataset. `Human' indicates the performance of the second annotator. `$DRIU^{*}$' represents our own implementation, which was required as a component of the proposed VGN. Comparison results are thankfully provided by the authors.}
	\label{fig:quan_res}
\end{figure*}

\textbf{Quantitative Evaluation}:
We compare the proposed VGN with the current SotA~\cite{maninis16,ganin14,fu16} as well as several conventional approaches~\cite{soares06,orlando14,becker13} on precision recall curves. The curve is obtained by computing multiple precision/recall pairs using multiple vessel probability thresholds. We also present the average precisions (AP) and the maximum F1 scores as summary measures. 

The precision recall curves for DRIVE/STARE are summarized in Fig.~\ref{fig:quan_res}. The proposed method shows comparable performance to the original DRIU~\cite{maninis16} for DRIVE and shows the best performance for the STARE dataset. In both datasets, the proposed method achieves the highest AP scores. We note that our implementation of the DRIU method, denoted as $DRIU^{*}$ in Fig.~\ref{fig:quan_res}, gives slightly different performances than the original. Compared to $DRIU^{*}$, which is the baseline for the proposed VGN, we can clearly see improved performance for both datasets. For the CA-XRA dataset, VGN scored an AP of 0.915 while DRIU scored 0.899, showing a relative improvement of 1.78\%.

\begin{figure*}[t]
	\centering
	\begin{minipage}{1\linewidth}
		\subfloat{\includegraphics[width = 0.1667\linewidth, height = 0.07\paperheight]{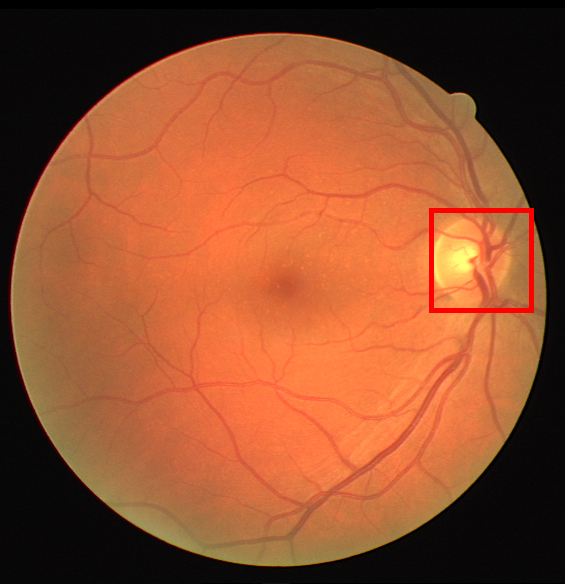}}
		\subfloat{\includegraphics[width = 0.1667\linewidth, height = 0.07\paperheight]{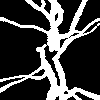}}
		\subfloat{\includegraphics[width = 0.1667\linewidth, height = 0.07\paperheight]{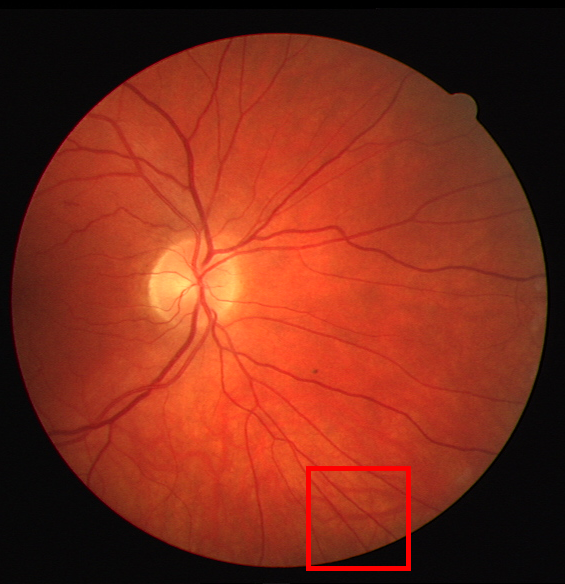}}
		\subfloat{\includegraphics[width = 0.1667\linewidth, height = 0.07\paperheight]{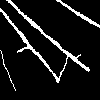}}
		\subfloat{\includegraphics[width = 0.1667\linewidth, height = 0.07\paperheight]{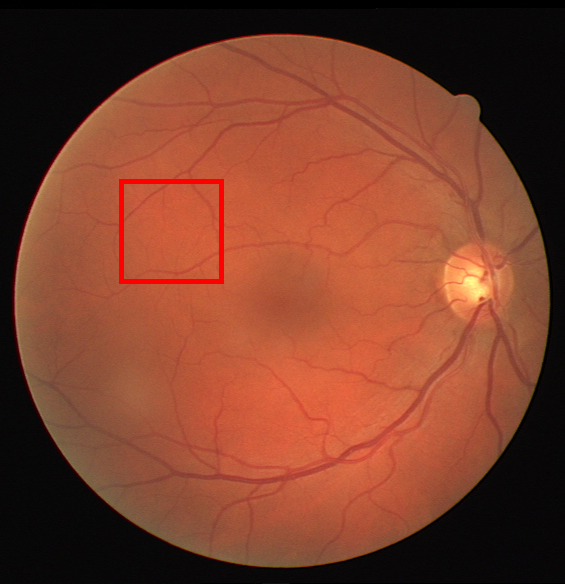}}
		\subfloat{\includegraphics[width = 0.1667\linewidth, height = 0.07\paperheight]{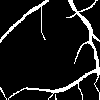}}
	\end{minipage}
	\\
	\vspace{-0.37cm}
	\begin{minipage}{1\linewidth}
		\subfloat{\includegraphics[width = 0.1667\linewidth, height = 0.07\paperheight]{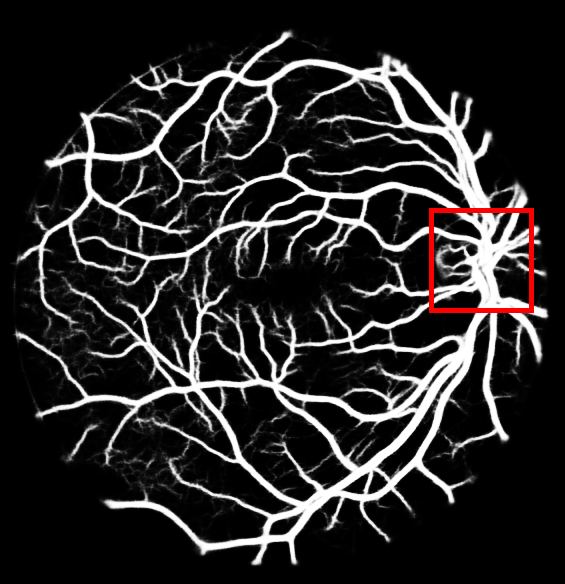}}
		\subfloat{\includegraphics[width = 0.1667\linewidth, height = 0.07\paperheight]{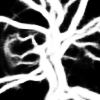}}
		\subfloat{\includegraphics[width = 0.1667\linewidth, height = 0.07\paperheight]{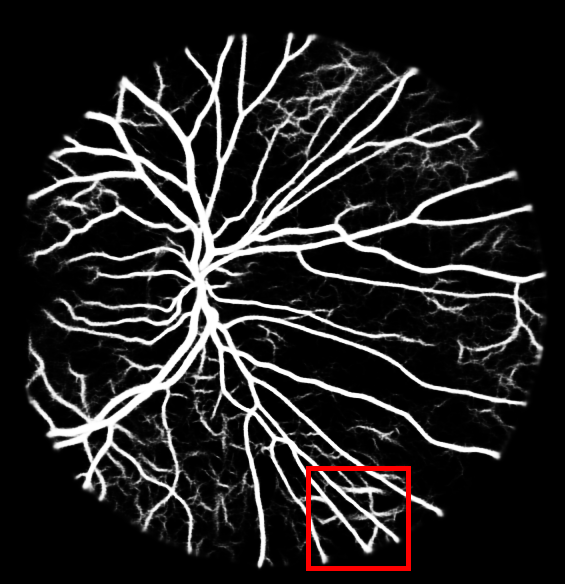}}
		\subfloat{\includegraphics[width = 0.1667\linewidth, height = 0.07\paperheight]{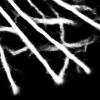}}
		\subfloat{\includegraphics[width = 0.1667\linewidth, height = 0.07\paperheight]{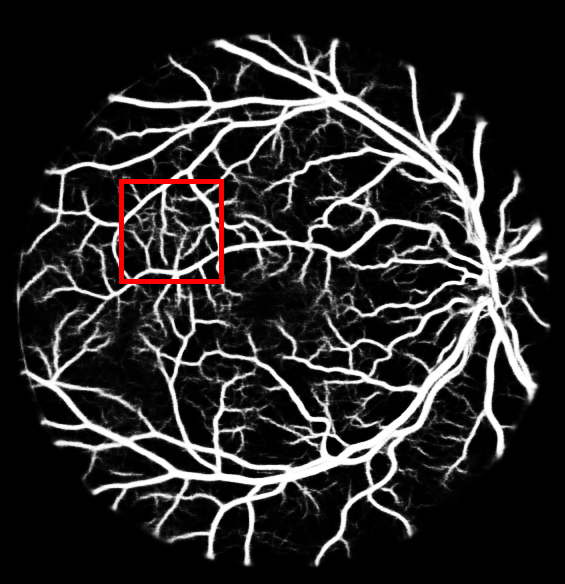}}
		\subfloat{\includegraphics[width = 0.1667\linewidth, height = 0.07\paperheight]{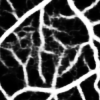}}
	\end{minipage}
	\\
	\vspace{-0.37cm}
	\begin{minipage}{1\linewidth}
		\subfloat{\includegraphics[width = 0.1667\linewidth, height = 0.07\paperheight]{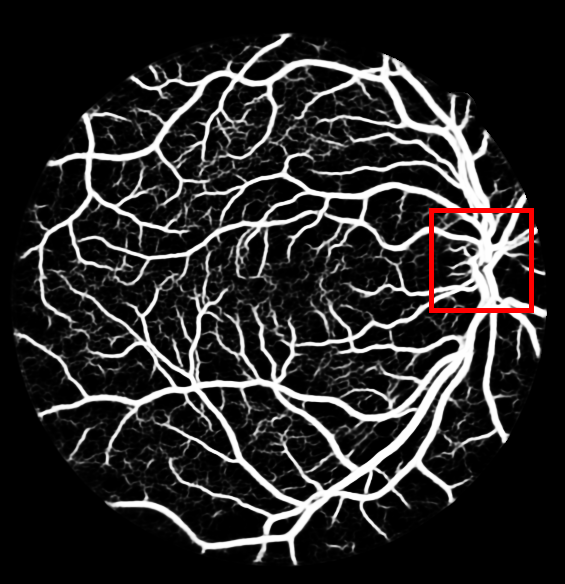}}
		\subfloat{\includegraphics[width = 0.1667\linewidth, height = 0.07\paperheight]{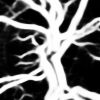}}
		\subfloat{\includegraphics[width = 0.1667\linewidth, height = 0.07\paperheight]{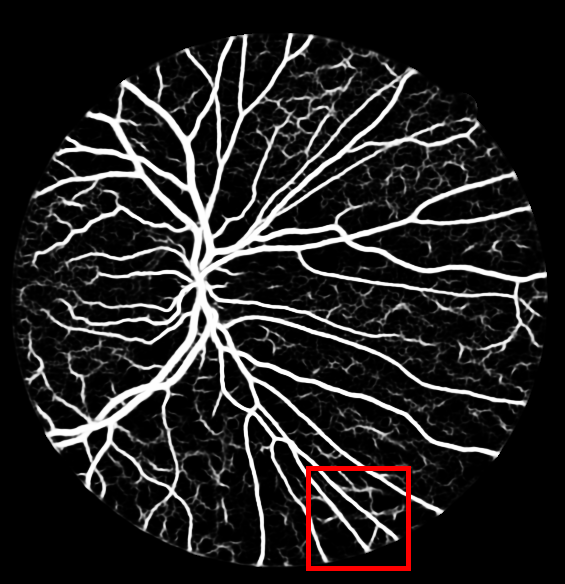}}
		\subfloat{\includegraphics[width = 0.1667\linewidth, height = 0.07\paperheight]{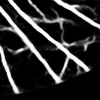}}
		\subfloat{\includegraphics[width = 0.1667\linewidth, height = 0.07\paperheight]{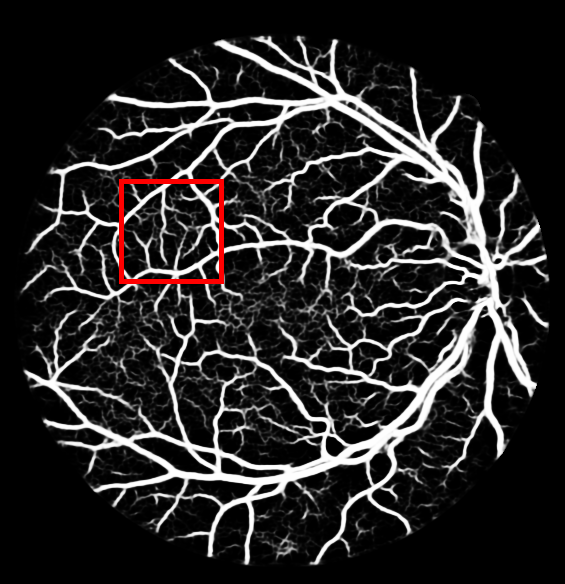}}
		\subfloat{\includegraphics[width = 0.1667\linewidth, height = 0.07\paperheight]{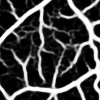}}
	\end{minipage}
	\\
	\vspace{-0.3cm}
	\begin{minipage}{1\linewidth}
		\subfloat{\includegraphics[width = 0.1667\linewidth, height = 0.07\paperheight]{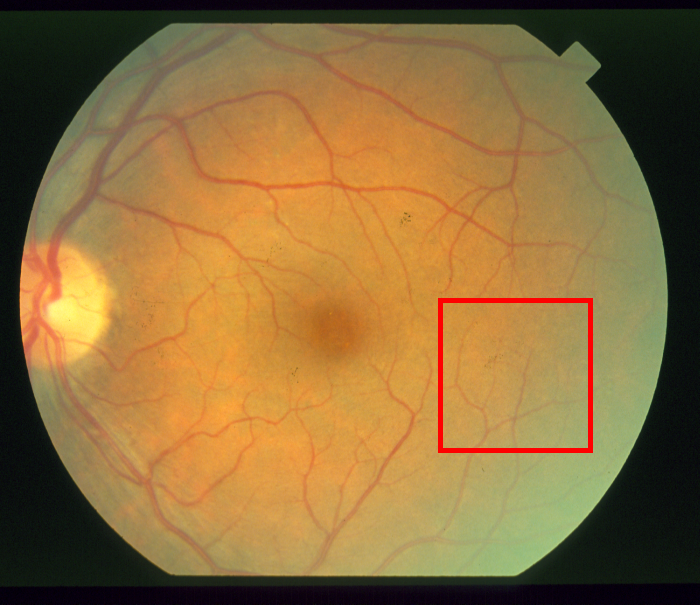}}
		\subfloat{\includegraphics[width = 0.1667\linewidth, height = 0.07\paperheight]{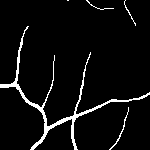}}
		\subfloat{\includegraphics[width = 0.1667\linewidth, height = 0.07\paperheight]{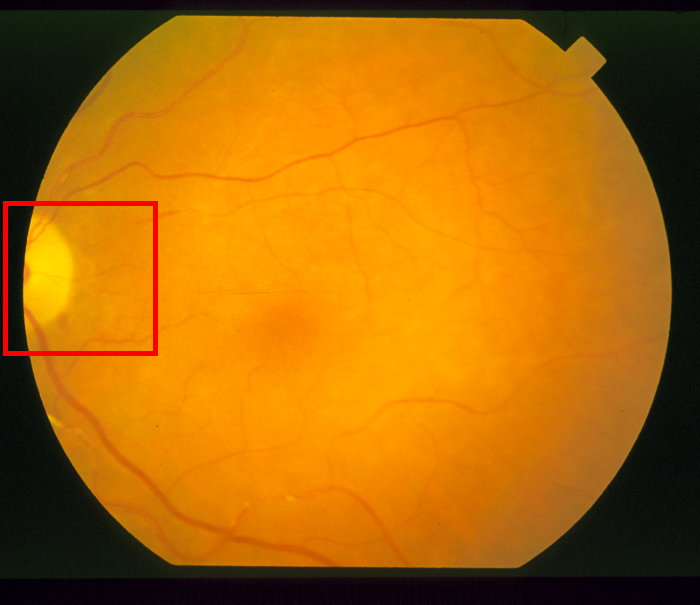}}
		\subfloat{\includegraphics[width = 0.1667\linewidth, height = 0.07\paperheight]{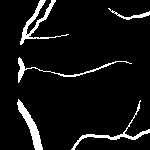}}
		\subfloat{\includegraphics[width = 0.1667\linewidth, height = 0.07\paperheight]{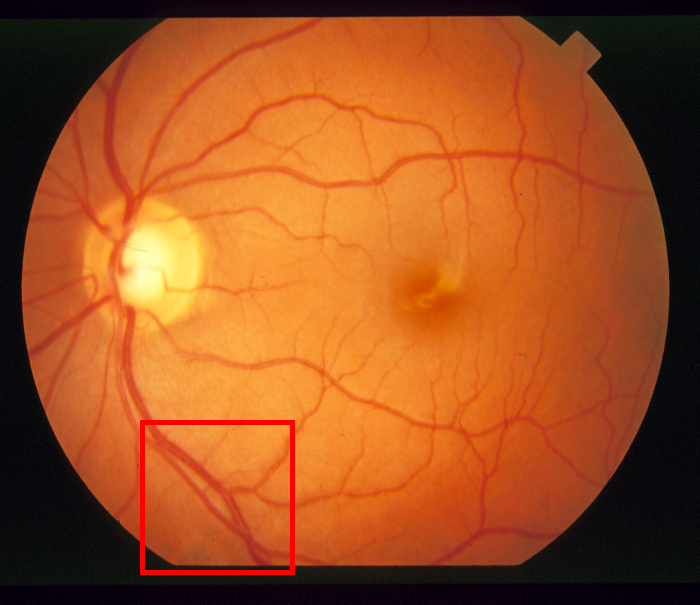}}
		\subfloat{\includegraphics[width = 0.1667\linewidth, height = 0.07\paperheight]{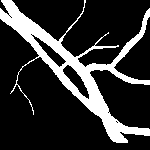}}
	\end{minipage}
	\\
	\vspace{-0.37cm}
	\begin{minipage}{1\linewidth}
		\subfloat{\includegraphics[width = 0.1667\linewidth, height = 0.07\paperheight]{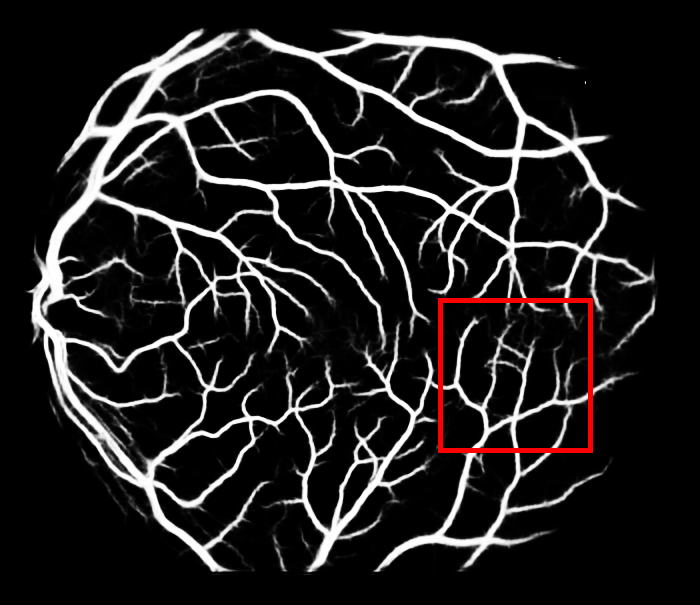}}
		\subfloat{\includegraphics[width = 0.1667\linewidth, height = 0.07\paperheight]{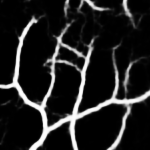}}
		\subfloat{\includegraphics[width = 0.1667\linewidth, height = 0.07\paperheight]{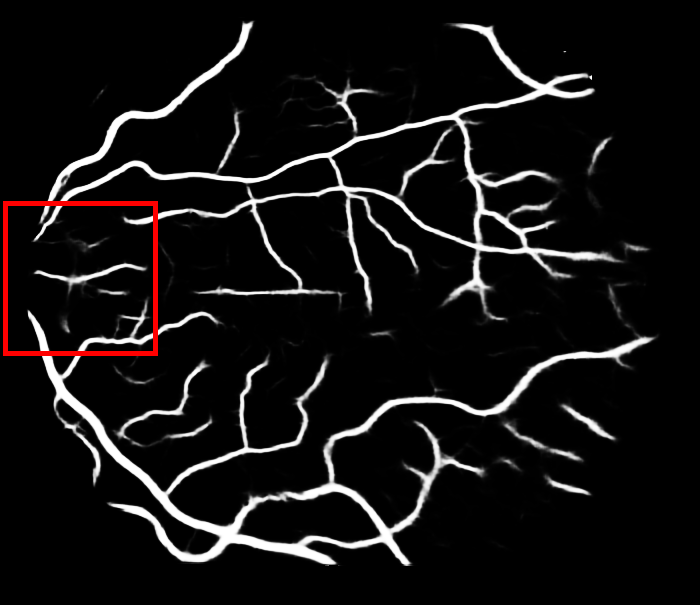}}
		\subfloat{\includegraphics[width = 0.1667\linewidth, height = 0.07\paperheight]{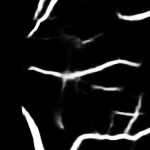}}
		\subfloat{\includegraphics[width = 0.1667\linewidth, height = 0.07\paperheight]{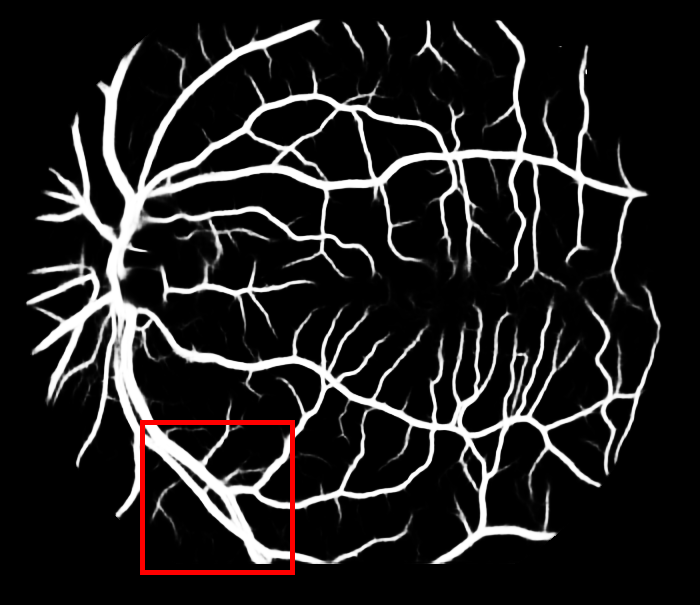}}
		\subfloat{\includegraphics[width = 0.1667\linewidth, height = 0.07\paperheight]{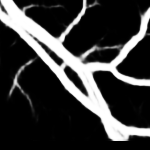}}
	\end{minipage}
	\\
	\vspace{-0.37cm}
	\begin{minipage}{1\linewidth}
		\subfloat{\includegraphics[width = 0.1667\linewidth, height = 0.07\paperheight]{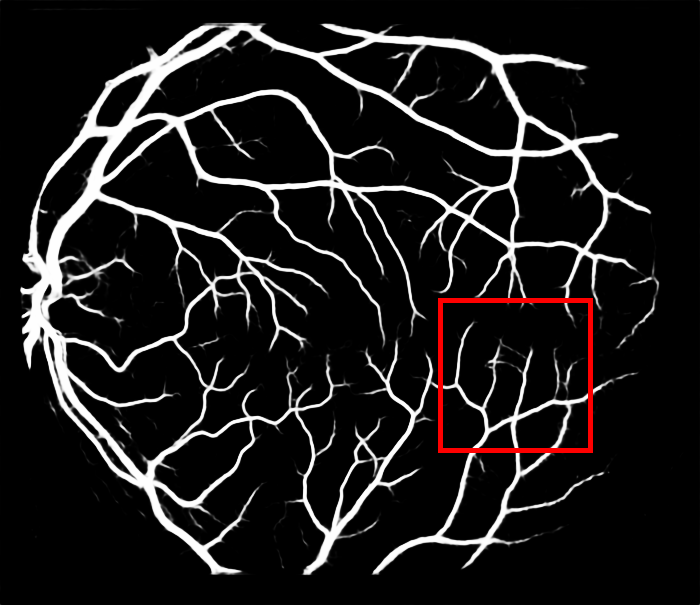}}
		\subfloat{\includegraphics[width = 0.1667\linewidth, height = 0.07\paperheight]{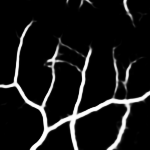}}
		\subfloat{\includegraphics[width = 0.1667\linewidth, height = 0.07\paperheight]{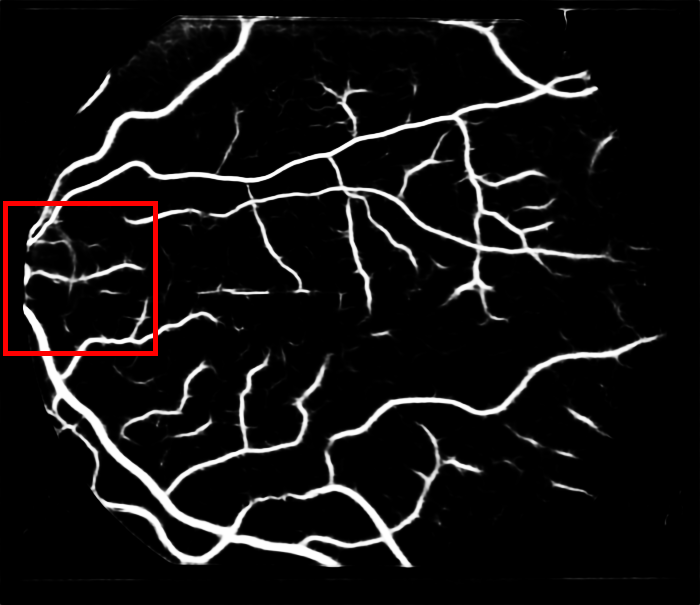}}
		\subfloat{\includegraphics[width = 0.1667\linewidth, height = 0.07\paperheight]{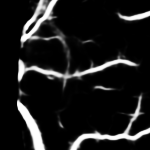}}
		\subfloat{\includegraphics[width = 0.1667\linewidth, height = 0.07\paperheight]{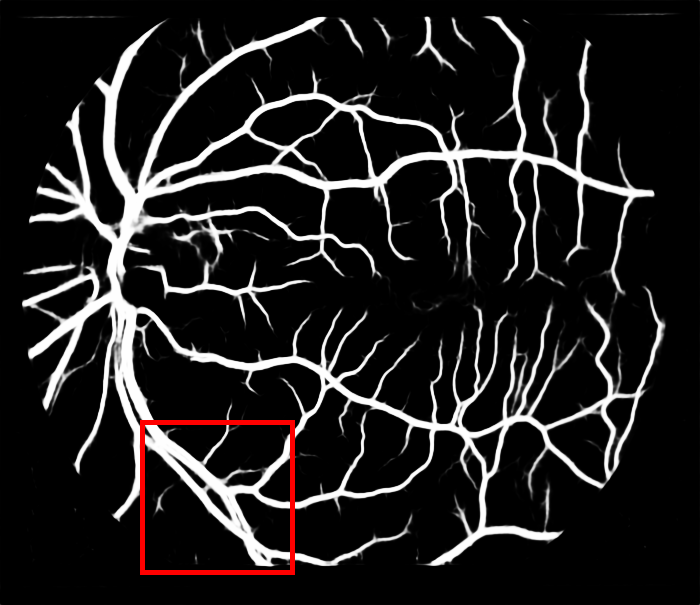}}
		\subfloat{\includegraphics[width = 0.1667\linewidth, height = 0.07\paperheight]{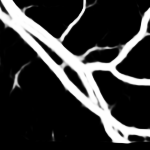}}
	\end{minipage}
	\\
	\vspace{-0.3cm}
	\begin{minipage}{1\linewidth}
		\subfloat{\includegraphics[width = 0.1667\linewidth, height = 0.07\paperheight]{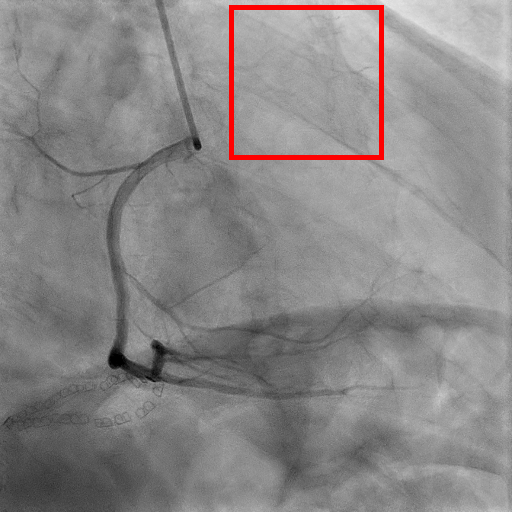}}
		\subfloat{\includegraphics[width = 0.1667\linewidth, height = 0.07\paperheight]{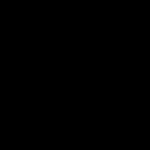}}
		\subfloat{\includegraphics[width = 0.1667\linewidth, height = 0.07\paperheight]{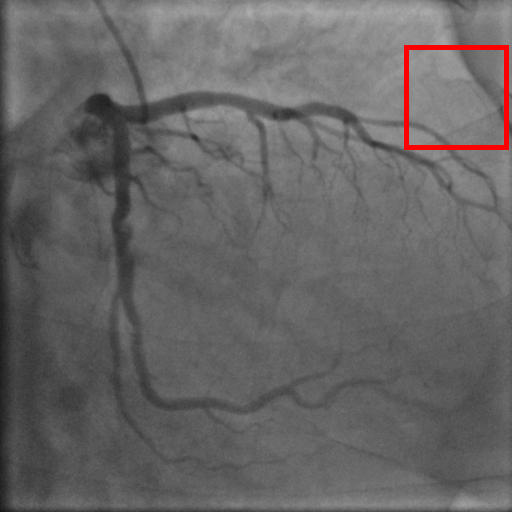}}
		\subfloat{\includegraphics[width = 0.1667\linewidth, height = 0.07\paperheight]{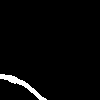}}
		\subfloat{\includegraphics[width = 0.1667\linewidth, height = 0.07\paperheight]{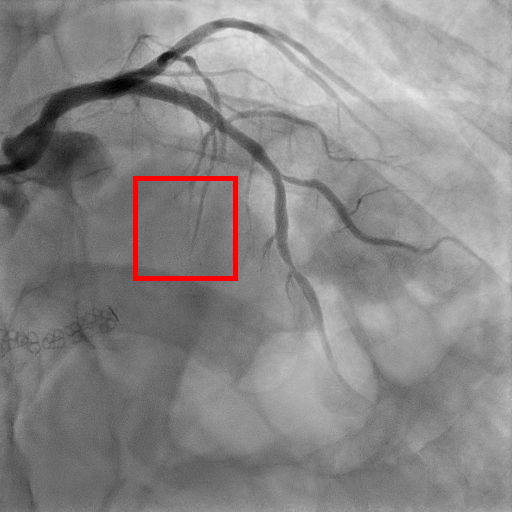}}
		\subfloat{\includegraphics[width = 0.1667\linewidth, height = 0.07\paperheight]{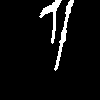}}
	\end{minipage}
	\\
	\vspace{-0.37cm}
	\begin{minipage}{1\linewidth}
		\subfloat{\includegraphics[width = 0.1667\linewidth, height = 0.07\paperheight]{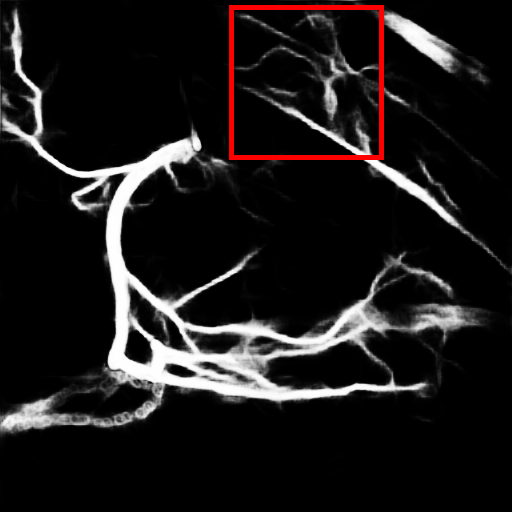}}
		\subfloat{\includegraphics[width = 0.1667\linewidth, height = 0.07\paperheight]{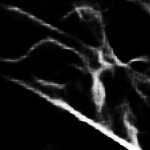}}
		\subfloat{\includegraphics[width = 0.1667\linewidth, height = 0.07\paperheight]{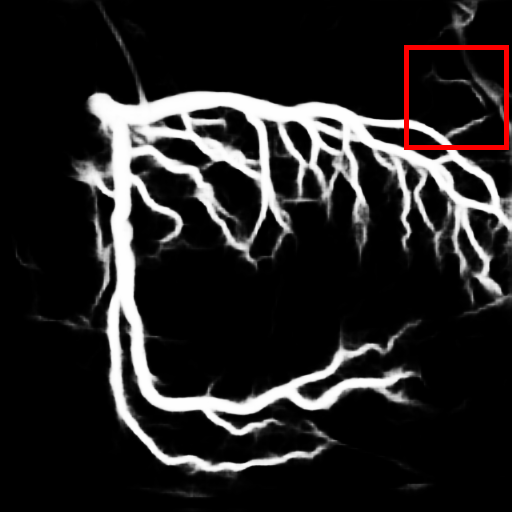}}
		\subfloat{\includegraphics[width = 0.1667\linewidth, height = 0.07\paperheight]{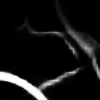}}
		\subfloat{\includegraphics[width = 0.1667\linewidth, height = 0.07\paperheight]{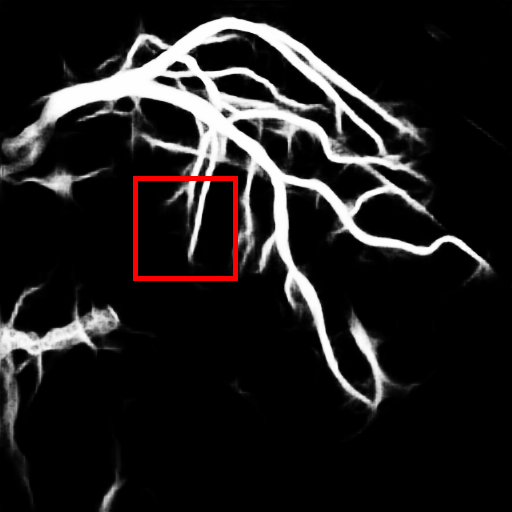}}
		\subfloat{\includegraphics[width = 0.1667\linewidth, height = 0.07\paperheight]{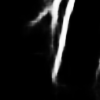}}
	\end{minipage}
	\\
	\vspace{-0.37cm}
	\begin{minipage}{1\linewidth}
		\subfloat{\includegraphics[width = 0.1667\linewidth, height = 0.07\paperheight]{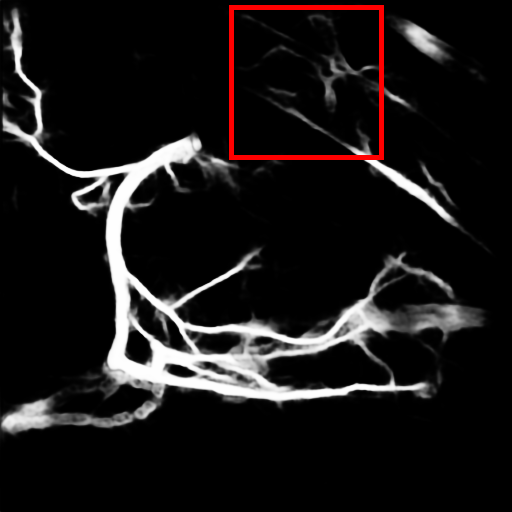}}
		\subfloat{\includegraphics[width = 0.1667\linewidth, height = 0.07\paperheight]{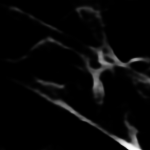}}
		\subfloat{\includegraphics[width = 0.1667\linewidth, height = 0.07\paperheight]{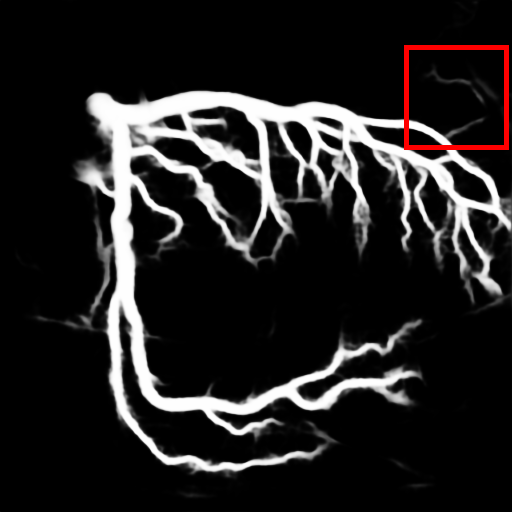}}
		\subfloat{\includegraphics[width = 0.1667\linewidth, height = 0.07\paperheight]{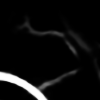}}
		\subfloat{\includegraphics[width = 0.1667\linewidth, height = 0.07\paperheight]{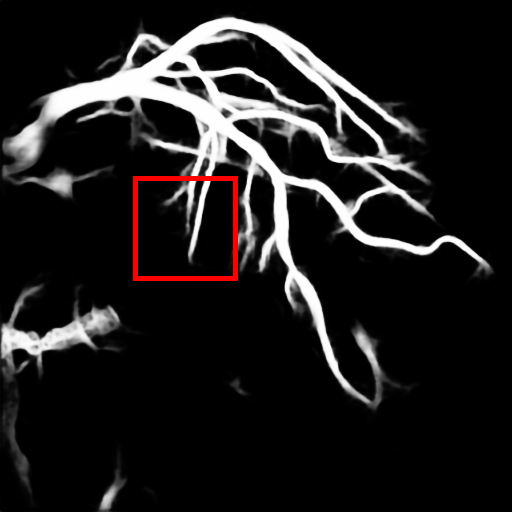}}
		\subfloat{\includegraphics[width = 0.1667\linewidth, height = 0.07\paperheight]{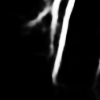}}
	\end{minipage}
	\caption{Qualitative results on the DRIVE, STARE, and CA-XRA dataset. Each of the three block rows represents each dataset in order, all of which with three representative sample results. Each of the six images for a single case represents an input image, result of \cite{maninis16}, result of VGC, zoomed GT, zoomed result of \cite{maninis16}, and zoomed result of VGC, from top-left to bottom-right. The results of \cite{maninis16} for DRIVE/STARE are those provided from the original authors, while our implementation results are shown for CA-XRA.}
	\label{fig:qual_res}
\end{figure*}

\textbf{Qualitative Evaluation}:
Fig.~\ref{fig:qual_res} shows qualitative results from each dataset. Compared to ~\cite{maninis16}, VGN reduces false positives, e.g., ribs in CA-XRA, and false negatives. It is interesting that very weak vessels in the first result of STARE are suppressed, rather than enhanced, by considering neighboring vessels. We also note that the proposed VGN seems to perform better for higher quality images such as the STARE dataset since the vessel graph structures become clearer.

\section{Conclusion}

We have proposed a novel CNN architecture that explicitly learns the graphical structure of vessel shape together with local appearance for vessel segmentation. Experiments show the effectiveness on three datasets about two different target organs. For future works, we plan to apply the proposed method to 3D imaging modalities such as the computed tomography angiography or extend it to use the temporal information of video data, e.g., fluoroscopic x-ray sequences.


\begin{thebibliography}{10}

\bibitem{soares06}
Soares, J.V.B., Leandro, J.J.G., Cesar, R.M., Jelinek, H.F., Cree, M.J.:
\newblock {Retinal vessel segmentation using the 2-D Gabor wavelet and
	supervised classification}.
\newblock IEEE T-MI \textbf{25}(9) (Sept 2006)
1214--1222

\bibitem{orlando14}
Orlando, J.I., Blaschko, Matthew", e.P., Hata, N., Barillot, C., Hornegger, J.,
Howe, R.:
\newblock {Learning Fully-Connected CRFs for Blood Vessel Segmentation in
	Retinal Images}.
\newblock In: MICCAI (2014)

\bibitem{shin16}
Shin, S.Y., Lee, S., Noh, K.J., Yun, I.D., Lee, K.M.:
\newblock {Extraction of Coronary Vessels in Fluoroscopic X-Ray Sequences Using
	Vessel Correspondence Optimization}.
\newblock In: MICCAI (2016)

\bibitem{becker13}
Becker, C., Rigamonti, R., Lepetit, V., Fua, P.:
\newblock {Supervised Feature Learning for Curvilinear Structure Segmentation}.
\newblock In: MICCAI (2013)

\bibitem{sironi15}
Sironi, A., Lepetit, V., Fua, P.:
\newblock {Projection onto the Manifold of Elongated Structures for Accurate
	Extraction}.
\newblock In: ICCV (2015)

\bibitem{ganin14}
Ganin, Y., Lempitsky, V.:
\newblock {$N^4$-Fields: Neural Network Nearest Neighbor Fields for Image
	Transforms}.
\newblock In: ACCV (2014)

\bibitem{fu16}
Fu, H., Xu, Y., Lin, S., Kee~Wong, D.W., Liu, J.:
\newblock {DeepVessel: Retinal Vessel Segmentation via Deep Learning and
	Conditional Random Field}.
\newblock In: MICCAI (2016)

\bibitem{maninis16}
Maninis, K.K., Pont-Tuset, J., Arbel{\'a}ez, P., Van~Gool, L.:
\newblock {Deep Retinal Image Understanding}.
\newblock In: MICCAI (2016)

\bibitem{kipf17}
Kipf, T.N., Welling, M.:
\newblock {Semi-Supervised Classification with Graph Convolutional Networks}.
\newblock In: ICLR (2017)

\bibitem{simonyan14}
Simonyan, K., Zisserman, A.:
\newblock {Very Deep Convolutional Networks for Large-Scale Image Recognition}.
\newblock CoRR \textbf{abs/1409.1556} (2014)

\bibitem{staal04}
Staal, J., Abramoff, M.D., Niemeijer, M., Viergever, M.A., van Ginneken, B.:
\newblock {Ridge-based vessel segmentation in color images of the retina}.
\newblock IEEE T-MI \textbf{23}(4) (April 2004)
501--509

\bibitem{hoover00}
Hoover, A.D., Kouznetsova, V., Goldbaum, M.:
\newblock {Locating blood vessels in retinal images by piecewise threshold
	probing of a matched filter response}.
\newblock IEEE T-MI \textbf{19}(3) (March 2000)
203--210
	
\end{thebibliography}

\end{document}